\documentclass[11pt]{article}

\usepackage[margin=1in]{geometry}
\usepackage[T1]{fontenc}
\usepackage{amsmath,amssymb,mathtools}
\usepackage{bm}
\usepackage{booktabs}
\usepackage{array}
\usepackage{graphicx}
\usepackage{xcolor}
\usepackage{caption}
\usepackage{microtype}
\usepackage[hidelinks]{hyperref}

\newcommand{\method}{ChronoStitch}
\newcommand{\norm}[1]{\left\lVert #1 \right\rVert}

\title{\bf \method: Training-Free Composition of Visual KV Memories for\\ Long-Horizon Temporal Reasoning}
\author{Santiram Tiwari, Nishant Sinha, and Kunal Kislay\\[3pt]
\normalsize KGraph AI Solutions Pvt.\ Ltd., Bangalore, India -- 560016}
\date{}

\begin{document}
\maketitle

\begin{abstract}
\noindent
Long-video question answering requires a model to preserve visual evidence over time without
repeatedly reprocessing the same video. A practical approach is to store the vision--language
model's internal key--value (KV) cache for each video chunk and retrieve that state at query
time. However, independently cached video chunks do not compose correctly: every chunk is
prefilled from local rotary position zero, so naive concatenation collides temporal phases and
removes the global order required for questions about what happened first, how often events
occurred, or what changed across the video. This paper presents \method, a training-free method
for composing independently stored visual KV memories. The method first re-bases stored
post-rotary keys onto a global three-axis multimodal RoPE coordinate system that preserves time,
height, and width structure. We show why a one-dimensional scalar re-indexing is geometrically
inconsistent for visual tokens because it turns spatial order within a frame into false temporal
displacement. We then address the residual content gap left by positional repair: later chunks
were originally encoded without attending to earlier chunks. \method{} therefore selectively
recomputes a small fraction of high-deviation later-chunk visual tokens while allowing them to
attend over the composed cache. On Qwen2.5-VL-3B and the temporal split of TempCompass,
\method{} outperforms naive composition and position-only variants, improving event-ordering
accuracy while running 3.3$\times$ faster than full joint re-prefilling.

\end{abstract}

\section{Introduction}
Vision--language models are increasingly expected to answer questions over long surveillance,
industrial, and operational videos. The difficulty is not only the number of frames, but the need
to preserve temporally ordered evidence for future questions whose content is unknown at ingest
time. Text captions are useful but lossy: a caption written before the question is asked may omit
precisely the detail needed later. Storing the model's KV cache offers a richer alternative
because it preserves internal visual activations rather than a fixed natural-language summary.

The mechanism is straightforward for one retrieved chunk. During transformer attention, each token
produces query, key, and value projections, and the keys and values of past tokens can be stored
for later attention. Rotary position embedding (RoPE) encodes relative order through rotations in
the attention space~\cite{roformer}. In multimodal models such as Qwen2-VL and Qwen2.5-VL, visual
tokens are not merely arranged as a text-like sequence; their positions are represented through
multimodal RoPE coordinates that combine temporal and spatial axes~\cite{qwen2vl,qwen25vl}. This
three-axis structure is central to video because tokens in the same frame should share a temporal
coordinate even when they occupy different image patches.

Existing work provides important ingredients but does not fully solve the composition problem
addressed here. Text-side KV reuse systems such as KVLink and CacheBlend precompute and combine
cached document states, and CacheBlend shows that selective recomputation can repair quality loss
when independently encoded text chunks are fused~\cite{kvlink,cacheblend}. Video-side systems such
as ReKV, StreamMem, StreamKV, HERMES, and DSCache store, retrieve, compress, or hierarchically
manage video KV caches for streaming and long-video settings~\cite{rekv,streammem,streamkv,hermes,dscache}.
Separately, VideoRoPE and related positional-encoding studies show that video RoPE requires
explicit spatio-temporal structure~\cite{videorope}. These lines of work motivate \method{} but
leave open a specific question: how should independently prefilled visual KV chunks be composed
under a model-native three-axis mRoPE, and how much cross-chunk attention must be repaired to
recover temporal reasoning?

\method{} addresses this question as a two-part, training-free procedure. First, it directly
rotates stored post-rotary keys from local chunk coordinates to global video coordinates,
preserving the model's time--height--width position geometry. Second, it selectively recomputes
only the most affected visual tokens from later chunks while allowing them to attend to earlier
chunks. The method is evaluated on controlled order-sensitivity probes and the temporal split of
TempCompass, a benchmark designed to test temporal perception in video LLMs~\cite{tempcompass}. The
central finding is that positions are necessary but not sufficient: mathematically correct
re-basing reconstructs the layer-0 positional geometry, but temporal question answering improves
only after selective cross-attention repair.

\section{Methodology}
Let a long video be divided into chunks $C_1, C_2, \ldots, C_N$. Each chunk is independently passed
through the vision--language model and stored as a KV cache containing layer-wise keys
and values, token coordinates, and layer-0 visual embeddings. The retained embeddings add modest storage overhead, but they avoid re-running the vision tower when selected tokens are repaired. At query time, a retrieval step may
return multiple chunks. The objective is to assemble their stored caches so the question can attend
over them as if the chunks had originally been prefilled as one continuous video. In ordinary
transformer attention, hidden states are projected into queries, keys, and values as follows.
\begin{equation}
Q = hW_Q, \quad K = hW_K, \quad V = hW_V, \quad
\mathrm{Attn}(Q,K,V) = \mathrm{softmax}\!\left(\frac{QK^{\top}}{\sqrt{d}}\right)V .
\end{equation}

For one-dimensional text, RoPE rotates token features according to a scalar position. In Qwen-style
multimodal RoPE, a visual token instead has a coordinate $p = (p^t, p^h, p^w)$, corresponding to
time, height, and width. Frequencies are assigned to one of these axes, and the rotary angle for
each frequency depends only on the associated coordinate. This can be summarized as follows, where
$\sigma$ selects the axis assigned to frequency $i$.
\begin{equation}
\alpha_i(p) = p_{\sigma(i)}\,\theta_i, \qquad f(x,p) = R\big(\alpha(p)\big)\,x, \qquad p = (p^t, p^h, p^w).
\end{equation}

The composition problem appears because each stored chunk was originally encoded at local
coordinates beginning near zero. Naively concatenating $\mathrm{KV}_{C_1}$, $\mathrm{KV}_{C_2}$, and
$\mathrm{KV}_{C_3}$ therefore makes unrelated chunks share overlapping rotary phases. \method{}
first fixes this by assigning every chunk a non-overlapping global coordinate interval and applying
a delta rotation to the stored post-rotary keys. Because planar rotations compose additively, the
key that was stored at local coordinate $p$ can be moved to target coordinate $p'$ without
recovering the original pre-rotary key $W_k x$ and without retraining the model. Values are not
rotated.
\begin{equation}
\Delta_i = \big(p'_{\sigma(i)} - p_{\sigma(i)}\big)\theta_i, \qquad
R(\Delta)\,f(W_k x, p) = f(W_k x, p').
\end{equation}

This re-basing must be three-axis rather than scalar. If visual tokens are flattened into
consecutive scalar positions, two different patches from the same frame receive different scalar
time indices even though their true temporal coordinate is identical. The resulting attention
phases falsely encode spatial patch order as temporal displacement. Three-axis re-basing avoids
this by shifting time, height, and width coordinates according to the model's own visual-token
layout, matching the geometry of a joint prefill at layer 0.

However, positional correction alone cannot reproduce a joint prefill at deeper layers. In a joint
prefill, a token in a later chunk can attend to earlier chunks while its hidden state is being
formed. In independent prefill, that causal cross-chunk interaction never occurred. The residual
content gap at layer $\ell$ can be written as follows.
\begin{equation}
\Gamma^{\ell} = \mathbb{E}_t \norm{h_t^{\ell,\mathrm{joint}} - h_t^{\ell,\mathrm{indep}}}_2 .
\end{equation}

\method{} repairs this gap by recomputing only a selected subset of later-chunk visual tokens.
Candidate tokens are ranked using an oracle-free proxy: after a shallow recomputation against the
composed cache, tokens whose layer-1 keys move the most are considered most affected by missing
cross-chunk attention. The score is defined below.
\begin{equation}
\delta_t = \norm{k_t^{1,\mathrm{recomp}} - k_t^{1,\mathrm{stored}}}_2 .
\end{equation}

The top $\rho$ fraction of later-chunk visual tokens is recomputed through all transformer layers
while attending causally to the full composed cache. Their newly computed keys and values are
scattered back into the stored cache; all other tokens remain unchanged. In the four-chunk
TempCompass setting used in our experiments, repairing $\rho=0.35$ of the later-chunk visual
tokens corresponds to refreshing approximately 25\% of the total composed visual cache. This makes
the procedure training-free and query-time efficient: most visual memory is reused, the vision tower
is skipped, and only the selected fraction is refreshed. The dominant cost terms compare as follows, where $V$
is the vision-tower cost, $T$ is the composed token length, $L$ is the number of layers, $c_a$ and
$c_f$ are attention and feed-forward constants, $c_r$ is the re-basing constant, and $\rho$ is the
repair fraction.
\begin{equation}
C_{\mathrm{joint}} \approx V + L\big(c_a T^2 + c_f T\big), \qquad
C_{\mathrm{Chrono}} \approx L c_r T + 2\big(c_a |C| T + c_f |C|\big) + L\big(c_a \rho T^2 + c_f \rho T\big).
\end{equation}

The theoretical speedup is therefore close to $1/\rho$ for the dominant attention term, with
additional gains because the vision tower is not rerun. In implementation, the procedure can be
summarized as: retrieve stored chunks, shift their three-axis coordinates to a global timeline,
delta-rotate stored keys, concatenate the cache, rank later-chunk visual tokens by the
key-deviation proxy, recompute the selected tokens with attention over the composed cache, and
return the repaired KV memory for answering the user query.

\section{Results and Discussion}
Experiments were conducted with \texttt{mlx-community/Qwen2.5-VL-3B-Instruct-4bit} using MLX on
Apple M-series hardware. The model has 36 layers, head dimension 128, 16 query heads, 2 key/value
heads, and mRoPE frequency sections of 16 temporal, 24 height, and 24 width frequencies. Video
chunks contained three frames, and the stored prefixes were vision-only. Five policies were
compared: full joint prefill, \method, three-axis re-basing only, one-dimensional scalar re-basing,
and naive concatenation. The joint prefill is not a deployable cached-memory method; it is used as
an upper bound because it processes the relevant chunks together from scratch.

\begin{table}[htbp]
\centering
\caption{Key reconstruction fidelity relative to full joint prefill.}
\small\begin{tabular}{c cc cc cc}
\toprule
Layer & Naive cos. & Naive rel.\ MSE & Scalar1D cos. & Scalar1D rel.\ MSE & 3-axis cos. & 3-axis rel.\ MSE \\
\midrule
0  & 0.9945 & $9.7\times10^{-3}$ & 0.9921 & $1.3\times10^{-2}$ & 1.0000 & $3.6\times10^{-8}$ \\
1  & 0.8677 & $1.7\times10^{-1}$ & 0.8176 & $2.7\times10^{-1}$ & 0.9998 & $2.2\times10^{-4}$ \\
9  & 0.6538 & $6.6\times10^{-1}$ & 0.5804 & $8.0\times10^{-1}$ & 0.9900 & $2.1\times10^{-2}$ \\
18 & 0.8368 & $2.8\times10^{-1}$ & 0.8020 & $3.4\times10^{-1}$ & 0.9735 & $3.3\times10^{-2}$ \\
27 & 0.9266 & $3.6\times10^{-2}$ & 0.9057 & $4.9\times10^{-2}$ & 0.9375 & $2.6\times10^{-2}$ \\
35 & 0.8133 & $3.6\times10^{-1}$ & 0.7740 & $4.3\times10^{-1}$ & 0.9551 & $8.4\times10^{-2}$ \\
\bottomrule
\end{tabular}
\captionsetup{font=small}
\caption*{\small\emph{At layer 0, three-axis re-basing reconstructs the joint-prefill keys to
machine precision, while the scalar collapse is worse than naive concatenation.}}
\end{table}

Table 1 confirms the mechanism. At layer 0, where keys depend only on token embeddings and
positional phases, three-axis re-basing has a relative error of $3.6\times10^{-8}$, effectively
matching the joint prefill. Scalar re-basing is not merely less precise; it is geometrically wrong
for video tokens and performs worse than the naive baseline in this diagnostic. The deeper-layer
decline in cosine similarity is expected because hidden states are no longer position-only: they
also reflect attention history. The independently stored chunks did not observe earlier chunks
during their original computation, so no rotation of keys alone can recreate the missing content.

\begin{table}[htbp]
\centering
\caption{Controlled order-sensitivity and selective repair.}
\small\begin{tabular}{lcc}
\toprule
Policy / repair fraction & Accuracy & Order-swap pairs flipped \\
\midrule
Joint prefill            & 100.0\% & 6/6 \\
Naive concatenation      & 50.0\%  & 0/6 \\
Scalar1D re-basing       & 58.3\%  & 1/6 \\
Three-axis re-basing only & 41.7\% & 0/6 \\
Repair $\rho = 0.15$     & 66.7\%  & -- \\
Repair $\rho = 0.35$     & 100.0\% & 6/6 \\
\bottomrule
\end{tabular}
\captionsetup{font=small}
\caption*{\small\emph{The control composes video chunks in A--B and B--A order and asks which
appears first. Correct temporal memory should flip its answer when the order is swapped. The number of flipped pairs for $\rho=0.15$ was not logged separately.}}
\end{table}

The controlled order-swap probe isolates the cross-attention gap. Even with correct positions, the
re-based cache fails to encode order because the later chunk was originally memorized without access
to the earlier chunk. Repairing 15\% of later-chunk tokens improves the result, and repairing
approximately 35\% recovers the joint ceiling on this probe. This supports the paper's main negative
result: positions are necessary, but not sufficient. Temporal coherence also requires a limited
amount of cross-chunk content repair.

\begin{table}[htbp]
\centering
\caption{TempCompass temporal split results, $N = 590$ multiple-choice questions.}
\small\begin{tabular}{lccc}
\toprule
Method & Overall & Event ordering & Attribute change \\
\midrule
Joint prefill ceiling     & 63.9\% & 60.9\% & 67.0\% \\
\method                   & 54.1\% & 54.0\% & 54.2\% \\
Three-axis re-basing only & 49.8\% & 46.7\% & 53.1\% \\
Scalar1D re-basing        & 49.5\% & 47.4\% & 51.7\% \\
Naive concatenation       & 49.3\% & 47.0\% & 51.7\% \\
\bottomrule
\end{tabular}
\captionsetup{font=small}
\caption*{\small\emph{\method{} improves over naive by 4.8 points overall and 7.0 points on event
ordering. The gains are largest where cross-chunk temporal reasoning is most necessary.}}
\end{table}

On the real temporal benchmark, \method{} is the best non-oracle method. The difference between
naive, Scalar1D, and position-only re-basing is small at the question-answering level, even though
the representational diagnostic strongly favors three-axis re-basing. This is an important
interpretation: three-axis re-basing provides the correct positional substrate, but the measurable
downstream improvement on the tested 3B reader comes primarily from selective repair. The method
recovers roughly one-third of the naive-to-joint gap overall and about half of the gap on event
ordering, while attribute-change questions show smaller gains because many such questions can be
answered from local evidence rather than long-range temporal interaction.

\begin{table}[htbp]
\centering
\caption{Query-time efficiency over a sample of 12 videos.}
\small\begin{tabular}{lccc}
\toprule
Method & Mean query time & Relative cost & Tokens recomputed \\
\midrule
Joint re-prefill & 2411 ms & 1.00$\times$ & 100\% plus vision tower \\
\method{} & 748 ms & 0.31$\times$ & approx. 25\% of composed cache \\
\bottomrule
\end{tabular}
\captionsetup{font=small}
\caption*{\small\emph{This corresponds to a 3.26$\times$ mean speedup and 3.31$\times$ median
speedup for \method{} over joint re-prefilling, with a measured range of 3.19--3.43$\times$.}}
\end{table}

The efficiency result is central to the practical value of the method. Full joint prefill is the
most faithful way to answer a query because the model reprocesses the relevant chunks together.
However, it pays the vision-tower and full-attention cost for every query. \method{} reuses the
existing visual memory, applies a cheap elementwise key rotation, and recomputes only a selected
fraction of tokens. This gives a measured 3.3$\times$ query-time speedup while recovering a substantial
portion of the temporal-ordering benefit of the joint ceiling.

The study also has limitations. The reader model is relatively small, and the joint ceiling on
TempCompass is only 63.9\%, which compresses the observable margin. The repair fraction was chosen
from a small control, and the optimal $\rho$ may vary with model scale, chunk length, or video
horizon. Efficiency is reported as wall-clock latency on one hardware configuration; a
hardware-independent FLOP and memory analysis would strengthen the claim. Finally, while the scalar
layout is mathematically and representationally defective, the present 3B-model benchmark does not
yet show a large downstream QA gap between scalar and three-axis re-basing without repair.

\section{Conclusion}
\method{} shows that independently stored visual KV memories can be composed for long-horizon
temporal reasoning, but only if two distinct errors are addressed. The first is a positional error:
independently cached chunks carry overlapping local mRoPE phases. This is corrected by a
training-free three-axis delta rotation applied directly to stored post-rotary keys. The second is
a content error: later chunks were encoded without attending to earlier chunks. This is addressed
by selectively recomputing high-deviation visual tokens while allowing them to attend over the
composed cache. The method is therefore not simply a cache concatenation scheme or a
position-reindexing trick; it is a two-stage memory-composition procedure that restores both
temporal geometry and a limited amount of missing cross-chunk attention. Experiments demonstrate
exact layer-0 key reconstruction, expose the insufficiency of positions alone, improve TempCompass
temporal reasoning over all non-oracle baselines, and reduce query-time cost by 3.3$\times$ compared
with full joint re-prefilling. These results support \method{} as a practical step toward coherent,
query-time video memory for long-duration visual AI systems.

\end{document}